\documentclass[sigconf]{acmart}

\renewcommand\footnotetextcopyrightpermission[1]{} 
\usepackage{graphicx}
\usepackage{xcolor}
\usepackage{amsmath}
\usepackage{subfigure}
\usepackage{colortbl}
\usepackage{subcaption}
\usepackage{algorithm}
\usepackage{algpseudocode}

\AtBeginDocument{%
  }

\begin{document}

\title{Can Synthetic Data Improve Symbolic Regression Extrapolation Performance?}


\author{Fitria Wulandari Ramlan}
\email{f.wulandari1@universityofgalway.ie}
\orcid{0000-0001-5415-2534}
\affiliation{%
  \institution{School of Computer Science, University of Galway}
  \city{Galway}
  \country{Ireland}
}

\author{Colm O'Riordan}
\email{colm.oriordan@universityofgalway.ie}
\orcid{0000-0003-0449-8224}
\affiliation{%
  \institution{School of Computer Science, University of Galway}
  \city{Galway}
  \country{Ireland}
}

\author{Gabriel Kronberger}
\email{gabriel.kronberger@fh-hagenberg.at}
\orcid{0000-0002-3012-3189}
\affiliation{%
  \institution{Heuristic and Evolutionary Algorithms Laboratory, University of Applied Sciences Upper Austria}
  \city{Hagenberg}
  \country{Austria}
}

\author{James McDermott}
\email{james.mcdermott@universityofgalway.ie}
\orcid{0000-0002-1402-6995}
\affiliation{%
  \institution{School of Computer Science, University of Galway}
  \city{Galway}
  \country{Ireland}
}


\begin{abstract}
  Many machine learning models perform well when making predictions within the training data range, but often struggle when required to extrapolate beyond it. Symbolic regression (SR) using genetic programming (GP) can generate flexible models but is prone to unreliable behaviour in extrapolation. This paper investigates whether adding synthetic data can help improve performance in such cases. We apply Kernel Density Estimation (KDE) to identify regions in the input space where the training data is sparse. Synthetic data is then generated in those regions using a knowledge distillation approach: a teacher model generates predictions on new input points, which are then used to train a student model. We evaluate this method across six benchmark datasets, using neural networks (NN), random forests (RF), and GP both as teacher models (to generate synthetic data) and as student models (trained on the augmented data). 
  Results show that GP models can often improve when trained on synthetic data, especially in extrapolation areas. However, the improvement depends on the dataset and teacher model used. The most important improvements are observed when synthetic data from GPe is used to train GPp in extrapolation regions. Changes in interpolation areas show only slight changes. We also observe heterogeneous errors, where model performance varies across different regions of the input space. Overall, this approach offers a practical solution for better extrapolation.
  \textbf{Note:} An earlier version of this work appeared in the GECCO 2025 Workshop on Symbolic Regression. This arXiv version corrects several parts of the original submission. In particular, we have revised the interpretation of extrapolation results in Sections~\ref{M1}-\ref{t-test} and updated the Conclusion to more accurately reflect the findings. Corrections include an updated interpretation of the red-green heatmap results, and a re-evaluation of the statistical significance analysis to ensure consistency with the performance trends shown in the heatmaps.
\end{abstract}





\maketitle

\begin{figure*}[t]
    \centering
    \includegraphics[width=0.9\linewidth]{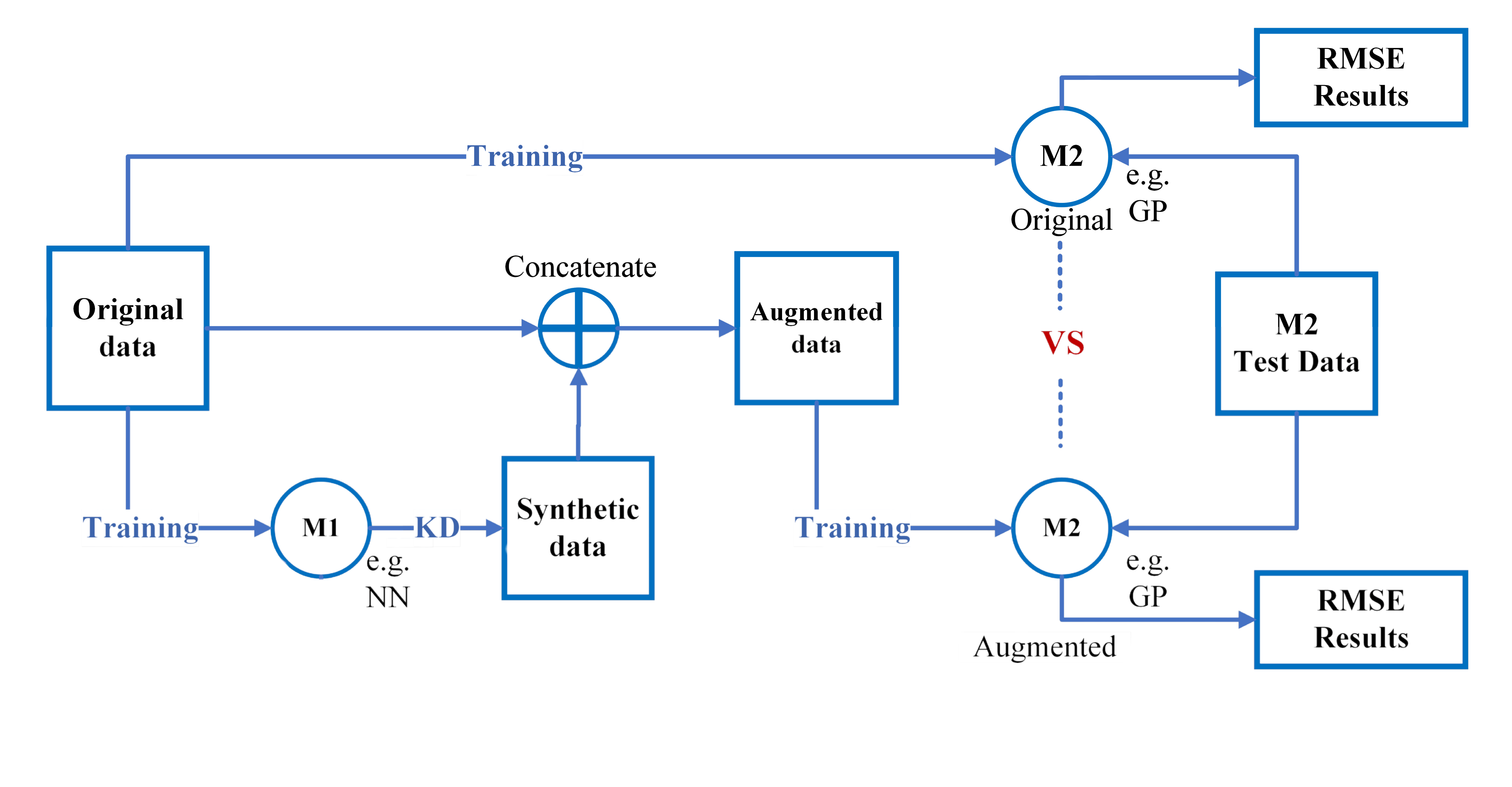}
    \caption{Knowledge Distillation Framework for Synthetic Data Augmentation. First, a teacher model (M1) is trained on original data and used to generate synthetic extrapolation data. This synthetic data is added to the original training set. Then, a student model (M2) is trained twice: once on the original data, and once on the augmented dataset. The performance difference between both M2 runs is measured to evaluate the benefit of synthetic extrapolation data.}
    \label{fig:hypothesis}
\end{figure*}

\section{Introduction}
Machine learning models are typically evaluated based on their ability to interpolate, i.e., predict outcomes within the range of their training data. However, many practical applications require extrapolation, where models are expected to make predictions in regions of the input space that are not covered during training~\cite{khoury2016interpolation}. In these cases, performance often gets worse, especially when models do not have access to domain-specific knowledge or constraints~\cite{cao2023extrapolation,angelis2023artificial}.

A major challenge in extrapolation is managing prediction error. Models tend to produce larger and more unpredictable errors in areas where they have not been trained~\cite{angelis2023artificial,wang2024extrapolation}. These unpredictable behaviours are often linked to heterogeneous errors (HE)~\cite{angelis2023artificial,hollmann2025accurate}, or systematic variations in error magnitude across the input space~\cite{dou2023machine,wang2024extrapolation}. This instability is similar to Runge's phenomenon~\cite{epperson1987runge}, where high-degree polynomial fits tend to oscillate or curve sharply near the edges of the data, leading to poor predictions in those regions. HE becomes more pronounced during extrapolation, where the training data provides insufficient coverage of the full problem domain.
To address this, our study investigates whether synthetic data augmentation can help reduce extrapolation error and HE. Rather than expanding the training dataset directly, we aim to expand the model's knowledge landscape. This is achieved by exposing the model to diverse inputs via artificially generated samples in regions with sparse training data. These extrapolation prone regions are identified using Kernel Density Estimation (KDE), a density-based method that estimates how training data is distributed in the input space. Areas with low density are treated as extrapolation zones, where errors are more likely to increase.

Another factor affecting extrapolation is model complexity. In symbolic regression (SR), complexity refers to the structure and mathematical form of the generated expressions, such as the number of mathematical operations, function types, and tree depth in a symbolic expression~\cite{bomarito2023automated,makke2024interpretable}. Simpler models are defined as those with lower structural complexity, such as fewer mathematical operations. While simpler models are generally easier to interpret, they may struggle to capture complex patterns in the data. However, more complex models can fit well but are prone to overfitting, especially in extrapolation scenarios~\cite{kronberger2022shape}. 

Recent studies show that synthetic data can improve model performance by increasing data diversity~\cite{choi2016retain,ramlan2023genetic,schmidt2009distilling}. In particular, earlier work by Schmidt and Lipson~\cite{schmidt2009distilling} generated synthetic data by sampling from smoothed spline fits before applying symbolic regression, which the conceptually similar strategy to ours. While earlier research has focused on the role of synthetic data in interpolation, its impact on extrapolation remains under-explored~\cite{cao2023extrapolation}. Specifically, there is limited understanding of how different models behave when trained with synthetic data, and how their extrapolation tendencies are affected by augmentation techniques. Moreover, the role of HE in extrapolation performance has not been widely investigated, despite its potential to affect model reliability in practical applications~\cite{cao2023extrapolation}. To support this investigation, we use KDE to differentiate between interpolation and extrapolation areas. KDE estimates how training data is distributed in the input space and helps identify areas where data is sparse. We treat these low-density areas as possible extrapolation regions. This allows us to observe how models behave in sections of the input space where there is limited exposure, and where they are more likely to make larger errors.

To address these gaps, we explore knowledge distillation (KD)~\cite{schmidt2009distilling}, where a pre-trained teacher model (M1) generates synthetic data to improve a student model's (M2) performance. Our primary hypothesis is that teacher models with stable extrapolation characteristics (i.e., NN, RF) can generate synthetic data that improve the extrapolation behaviour of student models (e.g., GP). For the teacher models, we apply Neural Networks (NN), Random Forests (RF), and GP. The student models are created using only GP expressions.
\begin{enumerate}
    \item Teacher Models (M1): NN, RF, GPp, and GPe are pre-trained on real data and then used to generate synthetic extrapolation data. Here, GPp refers to SR using GP with a heuristic-based model selection and GPe refers to selecting the model with the lowest training error.
    \item Student Model (M2): GP learns from both real and synthetic data to improve its extrapolation capability.
\end{enumerate}

Models like neural networks (NN) and random forests (RF) can be used as teacher models to generate synthetic data, though they show different behaviours in extrapolation. RF models consist of decision trees, which by design do not extrapolate and tend to produce constant predictions outside the range of the training data~\cite{fawagreh2014random}. This behaviour can help limit extreme values in extrapolated areas. In contrast, NN models typically overfit the training distribution and are known to perform poorly in extrapolation for regression tasks~\cite{garnelo2018conditional,ziyin2020neural}. Despite this, both RF and NN can still be valuable as teacher models in a KD framework, as their predictions typically smoother and more regular within or near the training domain. The goal is not to rely on their extrapolation ability, but rather to use them to generate informative and bounded synthetic targets in sparse regions of the input space. In contrast, GP models often show highly unpredictable or unstable behaviour in extrapolation due to their unconstrained symbolic expressions, which makes them suitable candidates for improvement via synthetic data augmentation.

This study evaluates the effect of synthetic data on extrapolation performance and provides the following contributions:
\begin{enumerate}
    \item A framework for generating synthetic data targeted at improving extrapolation in machine learning models.
    \item An experimental analysis of different teacher-student model configurations, evaluating the impact of synthetic data augmentation across various benchmark datasets.
    \item A study of heterogeneous errors (HE) in extrapolation regions, exploring how model reliability varies when synthetic data is introduced.
\end{enumerate}

\section{Literature Review}
Symbolic regression (SR) via GP is highly flexible but prone to overfitting~\cite{raymond2019genetic}, particularly with complex data. Unlike traditional regression techniques, GP searches for symbolic expressions that best fit the data, but its unconstrained nature often leads to extreme extrapolation predictions. While constrained GP methods~\cite{kronberger2022shape} improve generalisation by enforcing shape constraints, they remain limited in highly non-linear datasets.

One critical factor affecting extrapolation reliability is heterogeneous errors (HE) which refer to variability in model errors across different input regions. Extrapolation errors are often larger and more unpredictable than interpolation errors because the model lacks sufficient training data in those regions~\cite{langdon2022failed}. GP models, in particular, can generate extreme predictions outside the training domain, leading to highly inconsistent error patterns in extrapolation~\cite{kronberger2022shape,langdon2022failed}.

SR and GP have become powerful tools for generating interpretable models from data. However, they face unique challenges in extrapolation due to overfitting and the unconstrained search space in GP, which can lead to unstable predictions, especially with sparse or noisy data. To address these issues, Kronberger et al.~\cite{kronberger2022shape} proposed shape-constrained SR by incorporating domain-specific constraints such as monotonicity and convexity. These constraints help models follow known patterns for improving accuracy in extrapolation. Their study found that constraining the solution space enabled better generalisation outside the training set, especially with noise. Expanding this approach, Haider et al.~\cite{haider2023shape} proposed a multi-objective GP framework that optimises both accuracy and shape constraints using interval arithmetic. This method improves model handling of noisy data and improves extrapolation reliability by controlling behaviour beyond the training region. Shape constraints help prevent overfitting and reduce unpredictable predictions in unseen data. Another study~\cite{raghav2024interactive} highlighted the challenges standard GP models face during extrapolation and explored alternative strategies including transformed variables and synthetic data, though these methods remain less effective in high-dimensional problems.

While most methods for detecting extrapolation rely on distance measures or model uncertainty, density estimation offers another useful approach. Kernel Density Estimation (KDE) is a non-parametric method that estimates how the input data is distributed~\cite{pimentel2014reviewKDE,sugiyama2007covariateKDE}. In this study, we use KDE to find areas in the input space where data is sparse, which we define as extrapolation regions. Although KDE is a simple method, it has not been widely used in symbolic regression to separate interpolation and extrapolation. By applying KDE with different bandwidths, we can better identify these regions and evaluate how synthetic data affects errors and generalisation performance. Another related approach to synthetic data generation comes from oversampling techniques such as SMOTE (Synthetic Minority Over-sampling Technique)~\cite{alkhawaldeh2023challenges} and ADASYN (Adaptive Synthetic Sampling)~\cite{he2008adasyn}, which create artificial samples by interpolating between existing data points~\cite{brandt2021comparative}. While these methods have been widely used in classification tasks, their assumptions may not hold in regression settings. For example, they may generate unrealistic samples in continuous spaces or fail to represent true data variability. In addition, they can introduce noise, distort the original data distribution, or lead to redundancy issues that may negatively affect regression model performance~\cite{amin2016comparing,hasanin2019severely,mujahid2024data}. Given these concerns, alternative approaches such as knowledge distillation (KD) provide a more structured and reliable method for generating synthetic data for regression and extrapolation tasks.

A key consideration in synthetic data generation is its potential to reduce HE by providing more representative samples in extrapolation regions~\cite{yu2023dataset}. The controlled generation of synthetic data allows models to encounter more balanced distribution of inputs, potentially leading to more stable error patterns. By using KD, we aim to explore how teacher models impact the degree to which HE is reduced in extrapolation.

\begin{algorithm}[H]
\caption{KDE-Based Synthetic Data Generation using Teacher Model}
\label{alg:kde_synth}
\begin{algorithmic}[1]
\Require Training inputs $X_{\text{train}}$, targets $y_{\text{train}}$
\Require Teacher model $\mathcal{M}$, noise level $\varepsilon$, bandwidth $h$
\Ensure Synthetic dataset $(\hat{X}, \hat{y})$

\State \textbf{Standardise} $X_{\text{train}}$
\State Fit KDE with Gaussian kernel and bandwidth $h$ on $X_{\text{train}}$
\State Compute log-density scores for all samples in $X_{\text{train}}$
\State Identify extrapolation points: select samples below 10th percentile
\State Initialise empty lists: $\hat{X} \gets [\,]$, $\hat{y} \gets [\,]$

\For{$i = 1$ to $N_{\text{synth}}$}
    \State Randomly select $x_{\text{base}}$ from low-density samples
    \State Sample noise vector $\delta \sim \mathcal{N}(0, \varepsilon^2 I)$
    \State Compute synthetic input: $\hat{x} = x_{\text{base}} + \delta$
    \State Predict target: $\hat{y}_i = \mathcal{M}(\hat{x})$
    \State Append $\hat{x}$ to $\hat{X}$, and $\hat{y}_i$ to $\hat{y}$
\EndFor

\State \Return Synthetic dataset $(\hat{X}, \hat{y})$
\end{algorithmic}
\end{algorithm}

\section{Methodology and Experiments}

\subsection{Overview of Teacher-Student Model Distillation}

KD in this study is a modelling technique where a teacher model transfers knowledge to a student model by generating synthetic training data. We used KD to improve extrapolation in SR. A pre-trained teacher model (e.g., NN, RF, GP) generates synthetic data, which is then used to train a student model (e.g., GP).

The primary goal is to transfer extrapolation knowledge from stable teacher models (e.g., NN, RF) to improve the generalisation ability of GP in extrapolated regions. GP tends to produce extreme predictions outside the training domain, meaning outputs that deviate from the expected range due to uncontrolled symbolic expressions or overfitting to noise. We hypothesise that knowledge transfer through synthetic data can reduce HE and stabilise extrapolation behaviour.

\subsection{Synthetic Data Generation and Preprocessing}\label{parameter-selection}
To differentiate between interpolation and extrapolation regions in the input space, we apply KDE, a non-parametric method that estimates the probability density function (PDF) of the training data. KDE enables model-independent identification of low-density regions, which are treated as extrapolation zones where models may behave unpredictably due to sparse training coverage.

We fit KDE using a Gaussian kernel on the standardised training features. Samples in the bottom 10\% of the log-density scores are flagged as extrapolation points. Synthetic data is then generated in these regions using a two-step process:
\begin{enumerate}
    \item \textbf{Feature Noise Addition}: Gaussian noise with standard deviation $\varepsilon = 0.3$ is added to the selected extrapolation inputs to generate synthetic inputs $\hat{X}$. This ensure the augmented data remains local while introducing diversity.
    \item \textbf{Prediction via Teacher Model}: A pre-trained teacher model (e.g., NN, RF, or GP) is used to predict synthetic targets $\hat{y}$ for each synthetic input $\hat{X}$. This model-guided approach aligns with previous strategies that sample from smoother approximations to augment training data.
\end{enumerate}

This KDE-based augmentation strategy allows synthetic data to be injected into the sparsest parts of the input space, supporting extrapolation analysis and heterogeneous error (HE) reduction. Unlike distance-based heuristics, KDE provides a principled way to locate under-sampled regions across arbitrary data distributions. The KDE threshold (10\%), bandwidth (0.3), and noise level ($\varepsilon = 0.3$) are fixed across all datasets to ensure consistency. The number of synthetic points is approximately one-ninth the number of interpolation samples.

Table~\ref{datasetsimplified} summarises the number of interpolation (inside) samples and synthetic samples per dataset. The complete procedure is described in Algorithm~\ref{alg:kde_synth}.

\begin{table}[t]
\centering
\caption{Number of inside (interpolation) samples and synthetic samples per dataset. Synthetic samples are set as approx.\ one-ninth of the inside samples. KDE parameters are fixed across all datasets (threshold: 10\%, bandwidth: 0.3, noise: 0.3).}
\begin{tabular}{|l|r|r|}
\hline
\textbf{Dataset} & \textbf{Inside Samples} & \textbf{Synthetic Samples} \\
\hline
Gurson & 1671 & 200 \\
Pollen & 2700 & 300 \\
AF1 & 2700 & 300 \\
AF2 & 2700 & 300 \\
Red Wine & 1439 & 200 \\
White Wine & 2700 & 300 \\
\hline
\end{tabular}
\label{datasetsimplified}
\end{table}

\subsection{Models and Hyperparameters}
To perform the regression tasks, we use three types of models: neural networks (NN), random forest (RF), and genetic programming (GP), each chosen for their ability to capture nonlinear patterns while supporting model diversity for teacher-student distillation. The NN and RF models were implemented using the \texttt{scikit-learn} library~\cite{scikit-learn}. The GP model was implemented using the \texttt{PySR} symbolic regression framework~\cite{cranmer2023interpretable} with a multi-objective approach.

The NN models were structured as a two-layer feed-forward network. The first hidden layer had \textbf{150} neurons, while the second layer had \textbf{75} neurons, both using the \textbf{tanh} activation function. The model was trained using the Adam \textbf{optimiser} with a regularisation parameter (\textbf{alpha}) of \textbf{0.0002} and a learning rate of \textbf{0.01} for up to \textbf{180} iterations. The RF model consisted of \textbf{1000} decision trees, each with a maximum depth of \textbf{25}. The \textbf{max\_features} parameter was set to ``\textbf{auto}'', allowing the algorithm to determine the optimal number of features at each split. GP was implemented using the PySR library. PySR was run with a multi-objective optimisation setup, using \textbf{four island populations}, each containing \textbf{200 individuals}, evolved over \textbf{100 generations}. The function set included binary operators \texttt{[+, -, *, /]} and unary operators \texttt{(log, sin)}. To avoid errors from divide by zero or invalid expressions, PySR uses protected operations that return safe values instead of crashing. 

\subsection{Multi-objective GP}
SR models often face a trade-off between accuracy and complexity. In this study, we use a multi-objective GP to ensure that the generated expressions that are both accurate and interpretable. 
PySR uses a Pareto front-based selection to maintain a set of solutions that balance two competing objectives: minimising prediction error (e.g., MSE) and reducing expression complexity (e.g., the number of mathematical operations). To further support interpretability, we also set a maximum complexity limit, which ensures the final expressions to not grow too large or unwieldy. This combination helps prevent overfitting while keeping the symbolic models easy to understand.

\begin{enumerate}
    \item \textbf{GP with Heuristic Selection (GPp)}: This selection evaluates Pareto-optimal solutions where neither accuracy nor complexity can be improved without worsening the other. Among the Pareto-optimal candidates, the GPp model was selected using the scoring function implemented in~\cite{cranmer2023interpretable}, defined as:
    \begin{equation}\label{eqn:next_row_pysr}
        Score = -\log \frac{(loss_i / loss_{i-1})}{(complexity_i - complexity_{i-1})}
    \end{equation}
    This scoring method helps identify the point where further gains in accuracy require a large increase in complexity, ensuring a good trade-off between simplicity and performance. 
    
    \item \textbf{GP with Lowest Error (GPe)}: GPe is coming from the multi-objective GP. This approach focuses only on selecting the model with the lowest error on the training data, prioritising accuracy over complexity. It is useful in cases where high predictive performance is the main goal, even if the resulting models are more complex.
\end{enumerate}

\section{Results and Discussion}
\subsection{Performance Difference Analysis}
Our main hypothesis is that synthetic data generated by teacher models such as NN and RF can improve the extrapolation performance of GP. To evaluate this hypothesis, as shown in Figure~\ref{fig:hypothesis}, we follow a two-step process: (1) The student model (M2) is first trained using only the original training dataset. (2) The same model is retrained on an augmented dataset that includes both the original data and synthetic extrapolation points generated by the teacher model (M1). Both versions of the M2 model, before and after augmentation, are evaluated on the same test set. To evaluate whether the knowledge distillation process improves extrapolation, we compare the performance of the two M2 models using a percentage-based difference formula:
\begin{equation}\label{eqn:difference}
    \textbf{Performance Difference (\%)} = \frac{M2_{\text{original}} - M2_{\text{after M1}}}{M2_{\text{original}}} \times 100
\end{equation}

Tables~\ref{tab:all_results} presents the percentage change in RMSE when M2 is trained on synthetic data generated by M1. The rows represent the teacher model (M1), while the columns represent the student model (M2). Two matrices are shown for each dataset, one for interpolation and one for extrapolation. The values in the cells show the relative change in test RMSE compared to training the student model on original data alone. A positive percentage (green) means the RMSE disimproved, i.e.~the model became more accurate. A negative percentage (red) means the RMSE improved, i.e.~the model got worse.

\subsection{GP as Teacher (M1)}\label{M1}
When GP is used as teacher model (M1), that is when synthetic data is generated using a GP models and then used to train another model (whether a NN, RF, or another GP), the extrapolation results are mixed. See the {\bf rows} labelled GPp and GPe. In some cases, such as in the \textbf{Gurson} dataset, we see some performance improvements, especially when GP as teacher model is used to train GP or NN models. However, in other datasets like \textbf{AF1} and \textbf{AF2}, GP as teacher model shows consistent disimprovements across nearly all student models. For example, in \textbf{AF2}, using GPp or GPe as teacher model results in a disimprovement for RF or NN.

Similarly in \textbf{AF1}, GPp as teacher model reduces performance of NN by $-73.44\%$, but shows improvement of RF by $+12.83\%$.
The \textbf{Pollen} dataset shows a mixed pattern. GPp as teacher model shows large disimprovement for NN ($-114.26\%$) and RF ($-55.25\%$), even GPe ($-105.50\%$). However, GPe as teacher model leads to improvements for all student models. This is the only dataset where GP as teacher leads to consistent improvements across the board. In the \textbf{White Wine} dataset, GPp and GPe as teacher show improvements in NN, but disimprovements in RF. In \textbf{Red Wine}, GPp as teacher improves NN, but disimproves RF and GP.

Overall, these results show that using GP to generate synthetic data does not work well for helping other models improve their extrapolation.


\subsection{GP as Student (M2)}
On the other hand, when GP is used as student model (M2) in extrapolation, training it on synthetic data from other models (NN, RF, GPp or GPe) often shows some improvements performance. See the {\bf columns} labelled GPp and GPe. Across all six datasets, we observe more improvements than disimprovements in most cases. For example, in the \textbf{Gurson} dataset, all eight combinations where GP is M2 (trained on data from NN, RF, GPp, and GPe) lead to improvements in extrapolation. In \textbf{Pollen}, 6 out of 8 pairings show improved performance, 5 out of 8 in \textbf{AF2}, and 5 out of 8 in \textbf{White Wine}. In \textbf{Red Wine}, the result is balanced (4 out of 8 improvements), and in \textbf{AF1} has 3 out of 8 improvements.
This suggests that GP models generally \emph{benefit more often than not} from synthetic data. Compared to other models, GP shows more capable of using synthetic training samples to improve its extrapolation ability, even if the improvements are sometimes small.

There is also a particularly interesting pattern in every single dataset, when \textbf{GPe is used as teacher model (M1)} and \textbf{GPp is the student model (M2)}, the extrapolation performance always improves. This trend suggests that GPe generates synthetic data that is helpful for training GPp. A possible explanation is that GPe (the lowest error of GP) and GPp (the heuristic defined in PySR) represent the problem in different ways. GPe may uncover predictive patterns that GPp cannot find on its own, helping GPp learn more effectively when trained on that synthetic data.

In practice, since synthetic data augmentation is not guaranteed to be helpful and can sometimes disimprove performance, we need a strategy to take advantage of improvements while avoiding disimprovements. We notice that good performance on test interpolation almost always predicts good performance on test extrapolation. Therefore, a good strategy would be to train multiple GP models, one using the original data, and others using synthetic data from any models (e.g., NN, RF, GPp, or GPe). Then, use a validation set in the interpolation region (available by splitting the training data) to select either the original M2 or the one trained using augmented data. This strategy can help us avoid disimprovements while still offering a real chance to improve GP performance in extrapolation.

\begin{table}[]
    \centering
    \caption{Test interpolation and extrapolation results for all datasets (AF1, AF2, Red Wine, White Wine, Pollen, and Gurson), showing the relative percentage change in RMSE after training M2 (rows) on synthetic data generated by M1 (columns), as defined in Eqn~\ref{eqn:difference}. Negative values (in red) show performance disimprovement, while positive values (in green) show improvement. The values are calculated using the formula Eq.~\ref{eqn:difference}. The right-side panels focus on extrapolation results which are our main focus on this study. 
    }
    
    \label{tab:all_results}

    \begin{tabular}{c}
        {\includegraphics[width=0.9\linewidth]{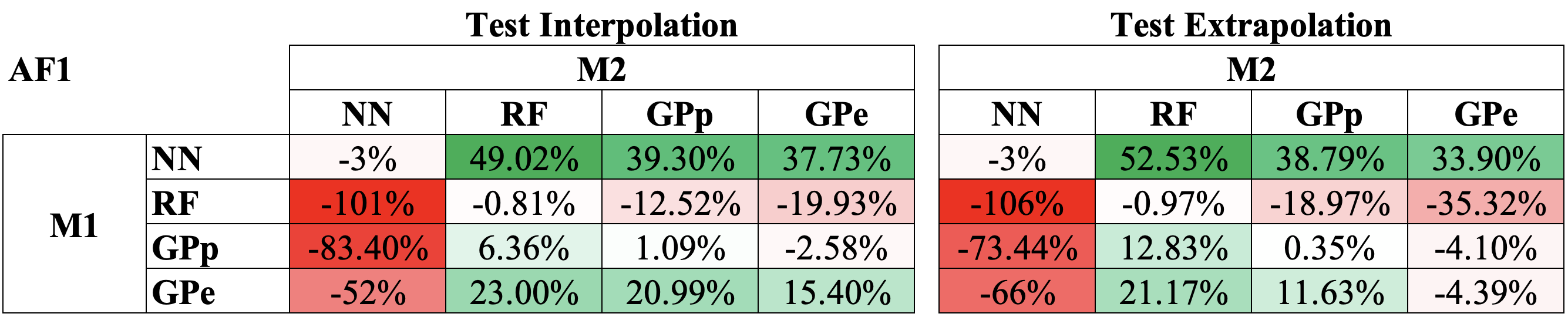} } \\ 
        {\includegraphics[width=0.9\linewidth]{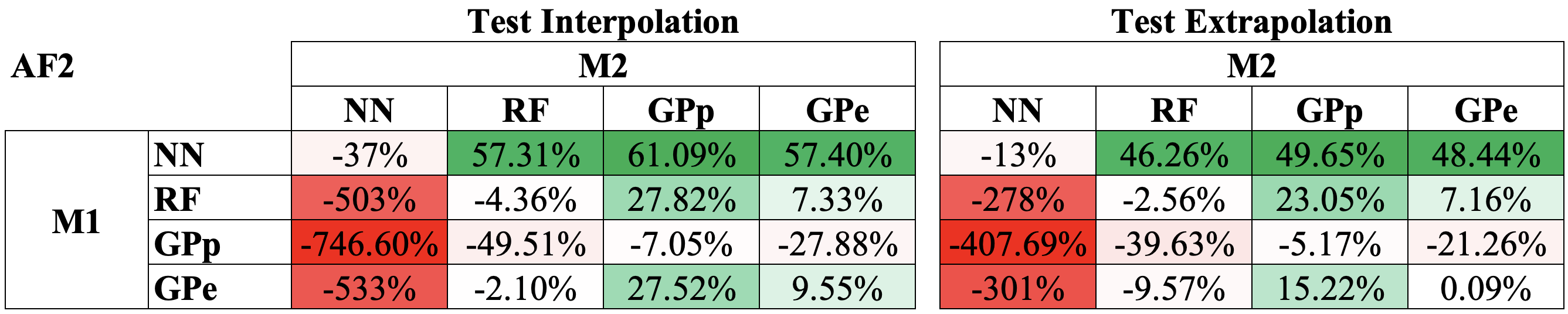} } \\ 
        {\includegraphics[width=0.9\linewidth]{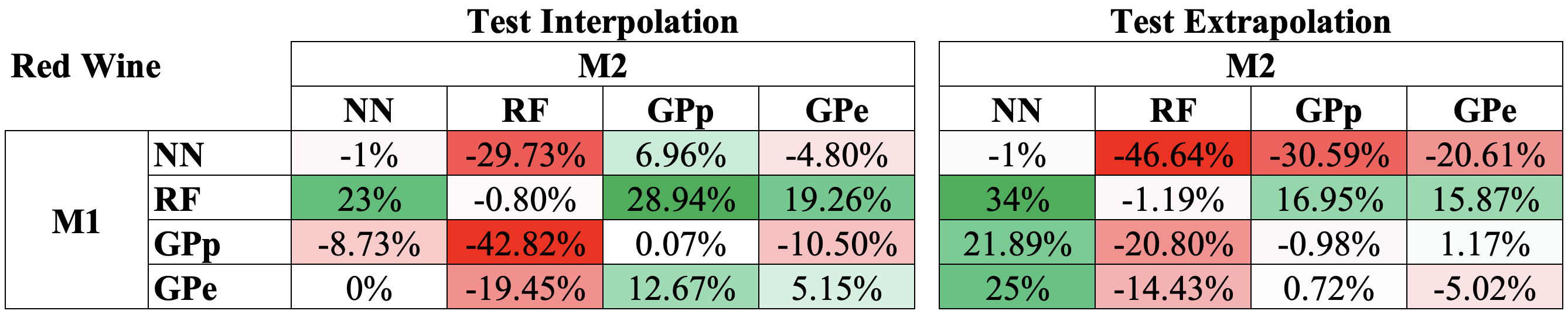} } \\ 
        {\includegraphics[width=0.9\linewidth]{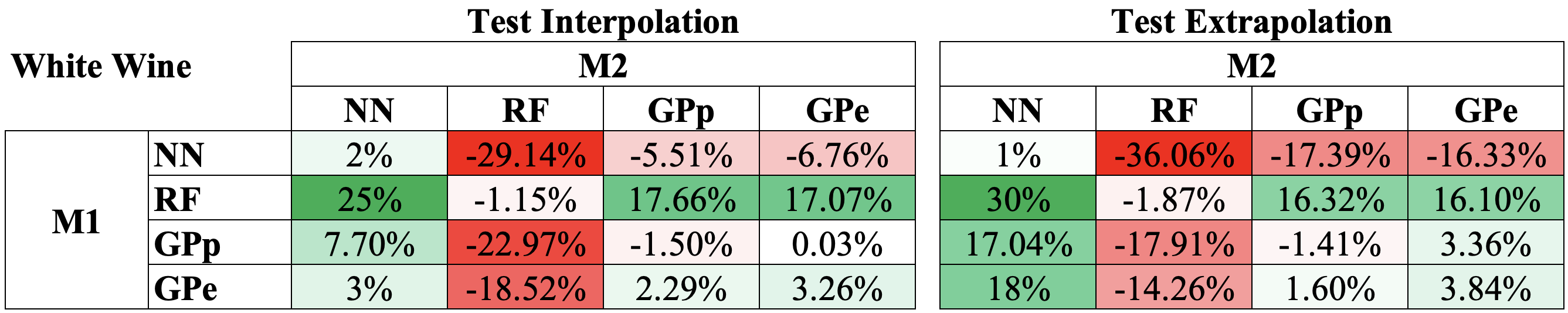} } \\ 
        {\includegraphics[width=0.9\linewidth]{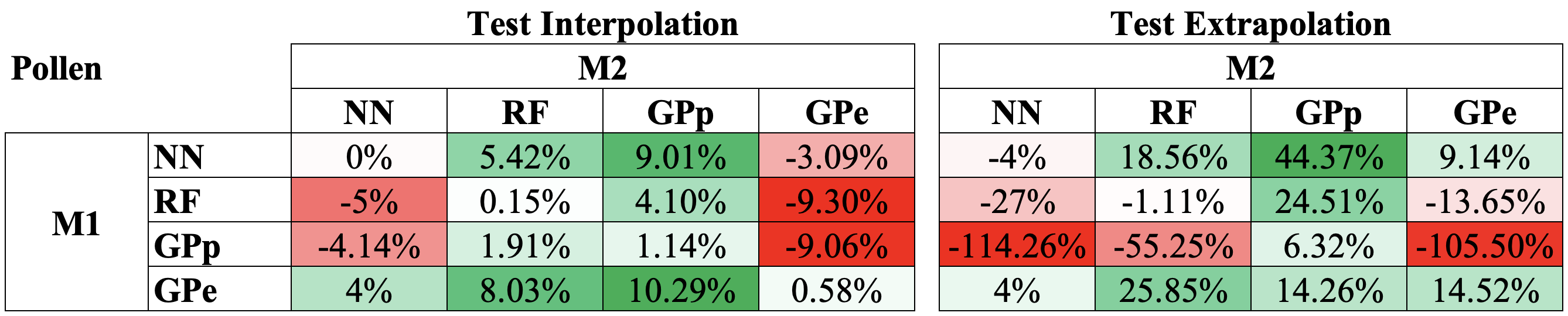} } \\ 
        {\includegraphics[width=0.9\linewidth]{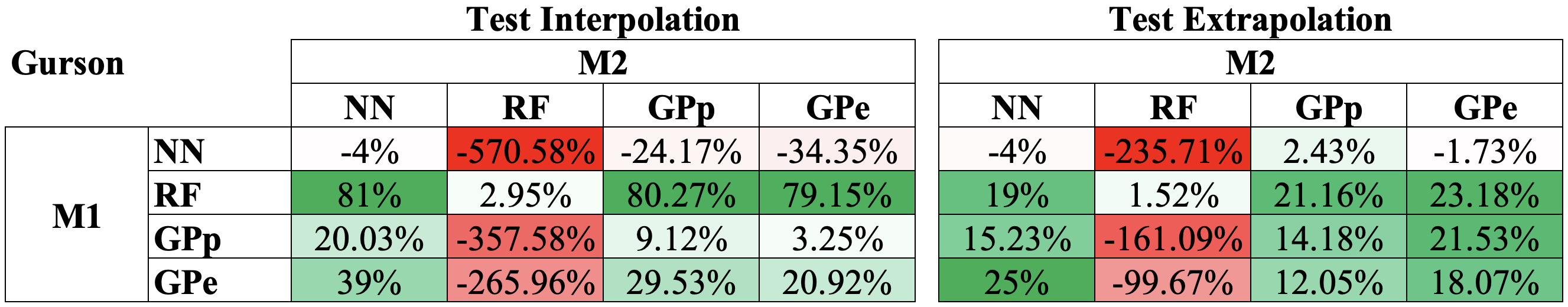} } \\ 
    \end{tabular}
\end{table}

\begin{figure}[]
    \centering
    \includegraphics[width=0.9\linewidth]{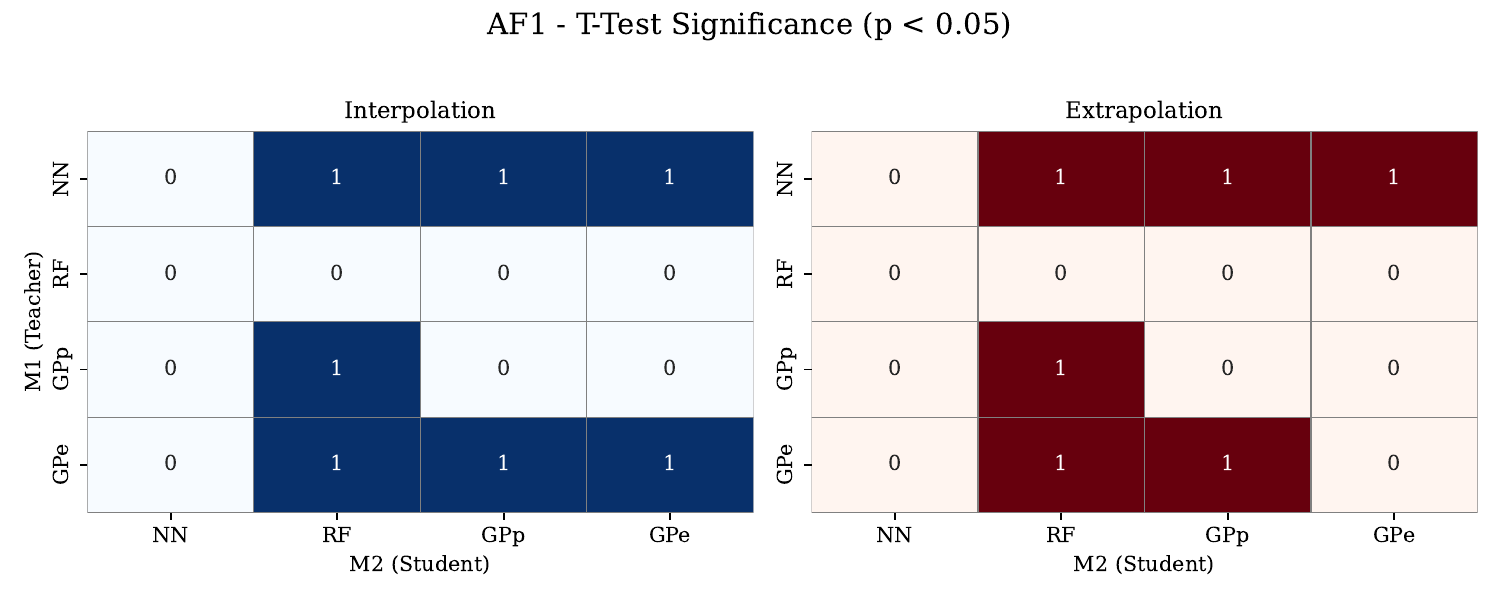}
    \includegraphics[width=0.9\linewidth]{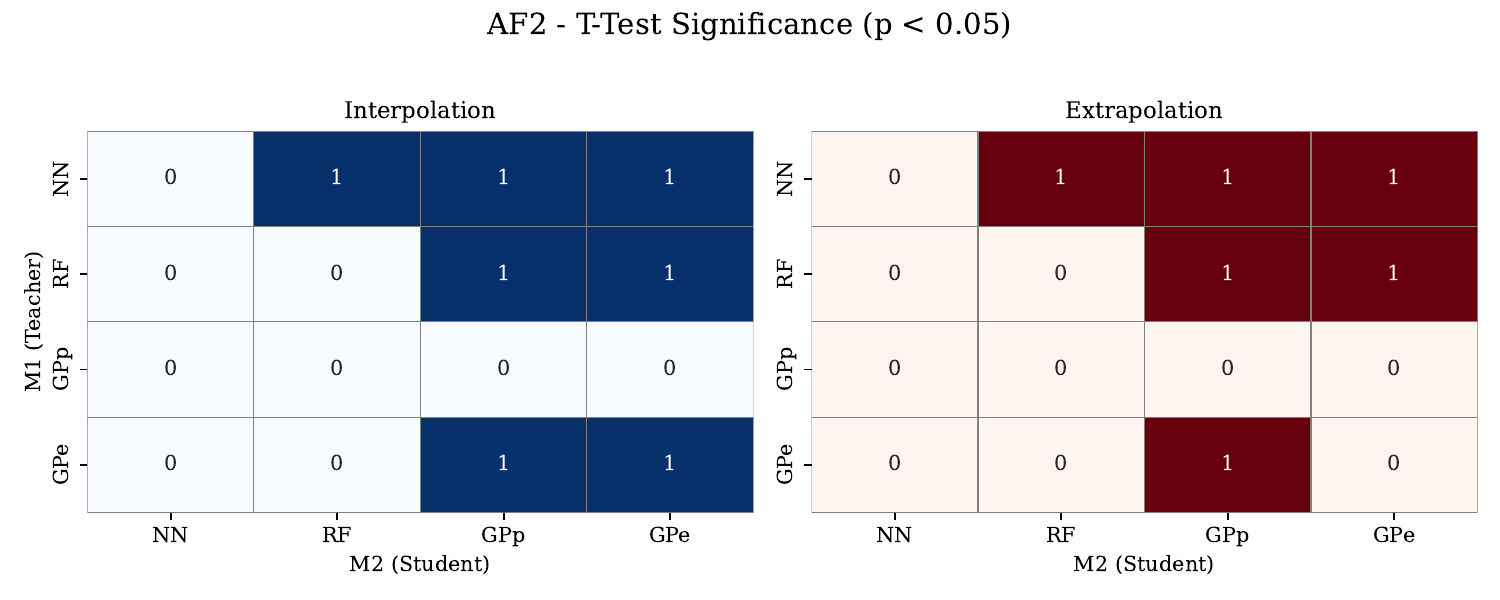}
    \includegraphics[width=0.9\linewidth]{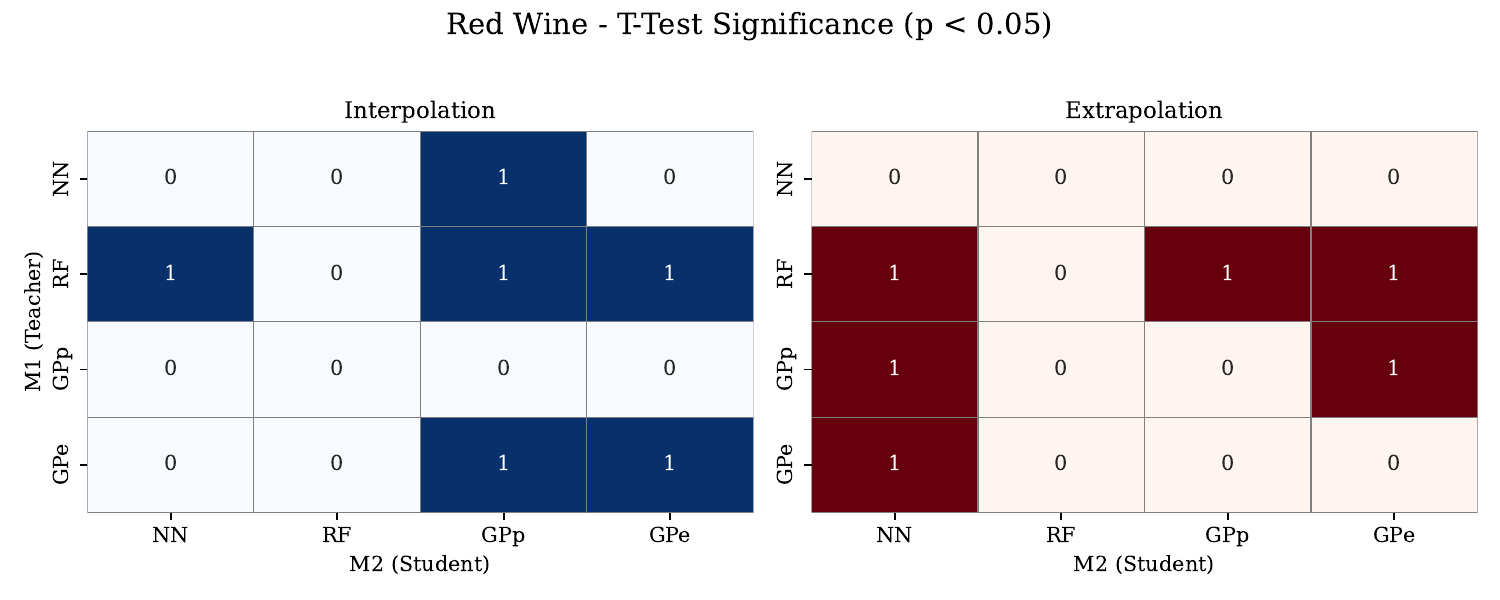}
    \includegraphics[width=0.9\linewidth]{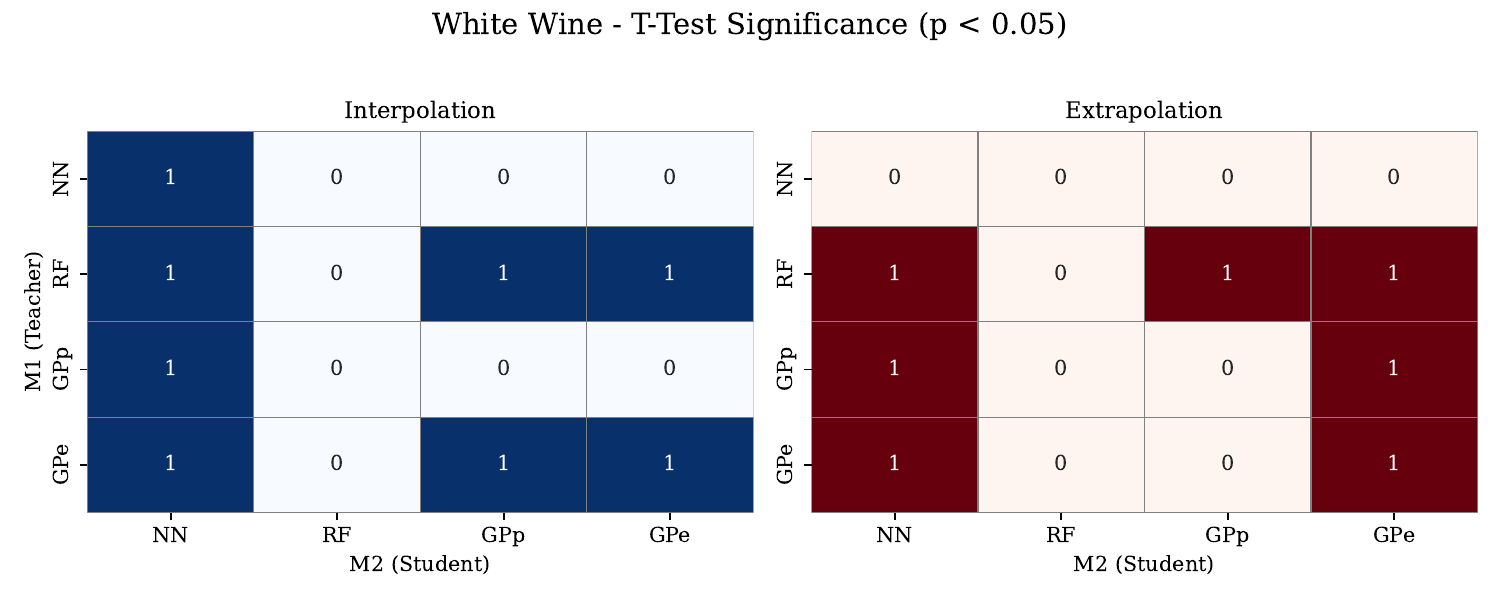}
    \includegraphics[width=0.9\linewidth]{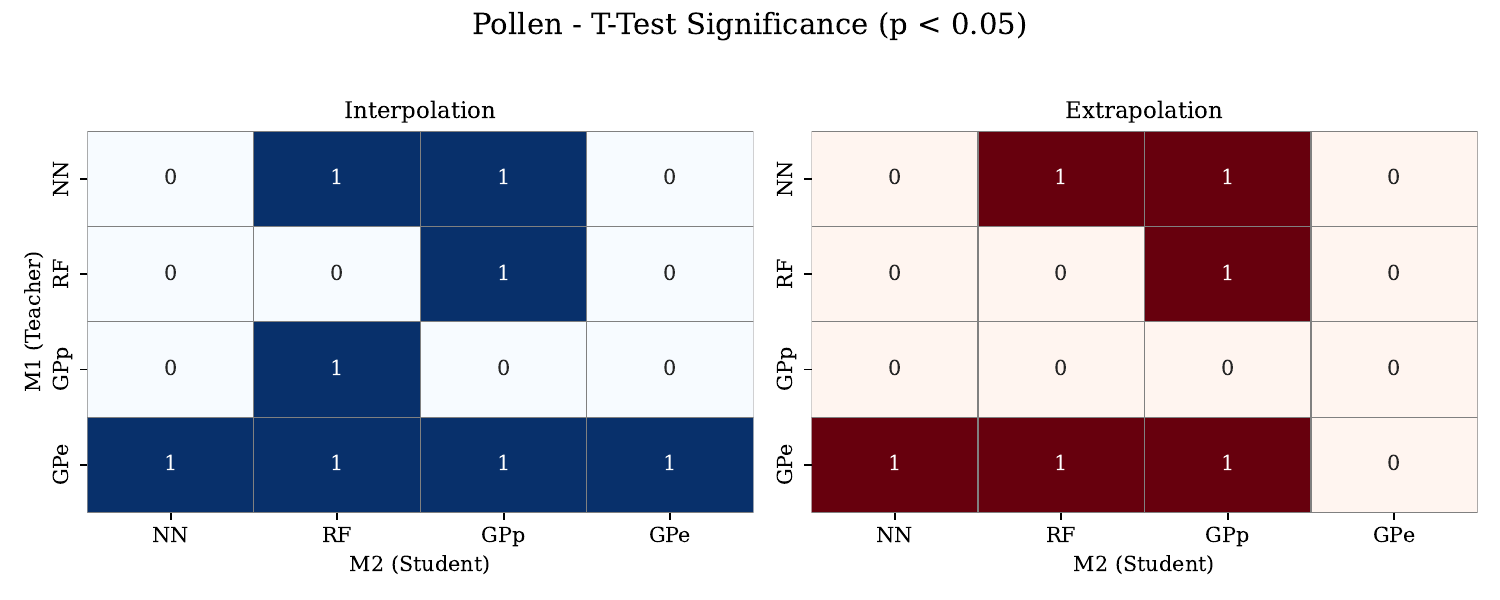}
    \includegraphics[width=0.9\linewidth]{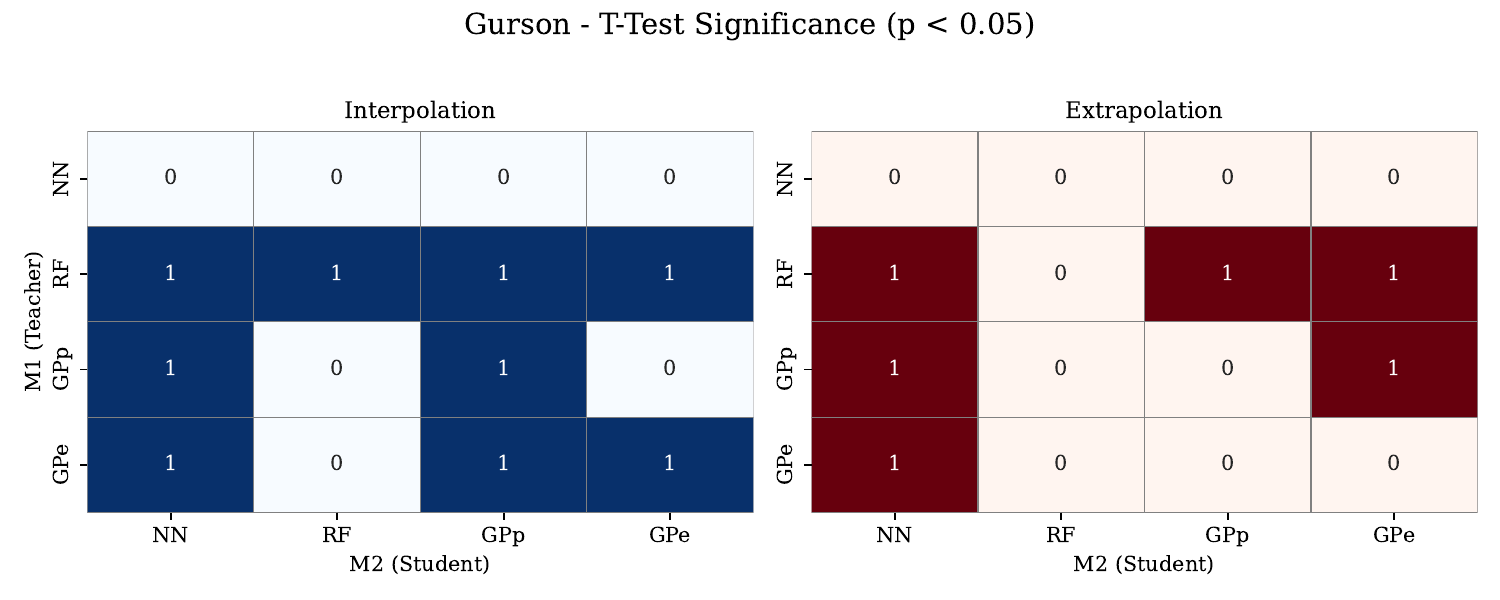}
    \caption{Significance heatmaps ($1 = p< 0.05$) for interpolation (left) and extrapolation (right) across all datasets. Each cell indicates whether the performance difference between the baseline model (M2 Alone) and the augmented model (M2 after M1) is statistically significant. A value of 1 (dark colour) means the performance difference is statistically significant at $p < 0.05$, and 0 (light colour) means the difference is not statistically significant.}
    \label{fig:significance-heatmap}
\end{figure}

\begin{figure}[]
    \centering
    \includegraphics[width=0.9\linewidth]{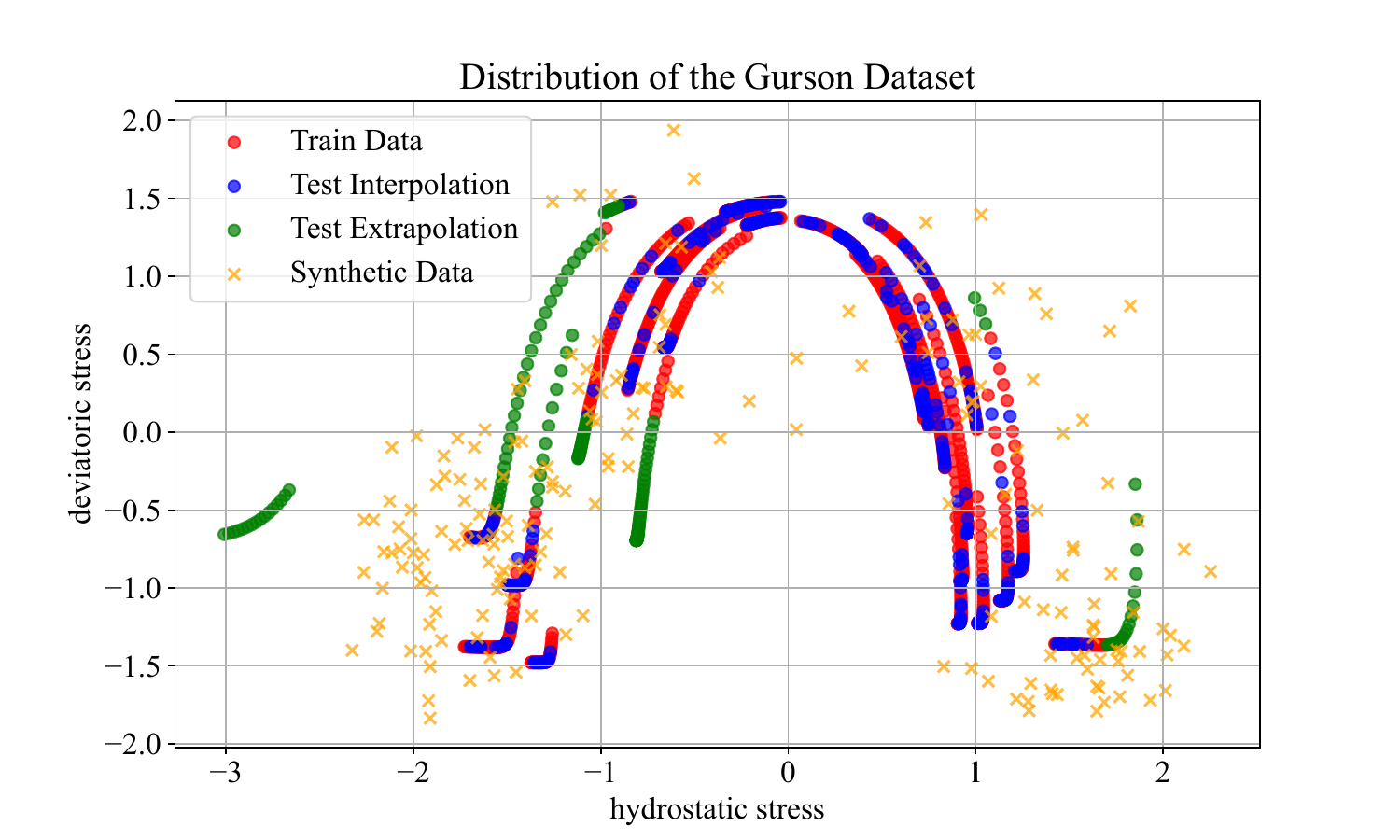}
    \caption{Data distribution in the Gurson dataset (Run 1), plotted for $X_1$ and $X_2$ axes. The plot shows the locations of the training data, test interpolation points, test extrapolation points, and synthetic data. Extrapolation points lie far outside the training area, while synthetic data aims to help models generalise better in these outers areas.}
    \label{fig:gurson_distribution}
\end{figure}

\begin{figure}[]
    \centering
    \includegraphics[width=0.9\linewidth]{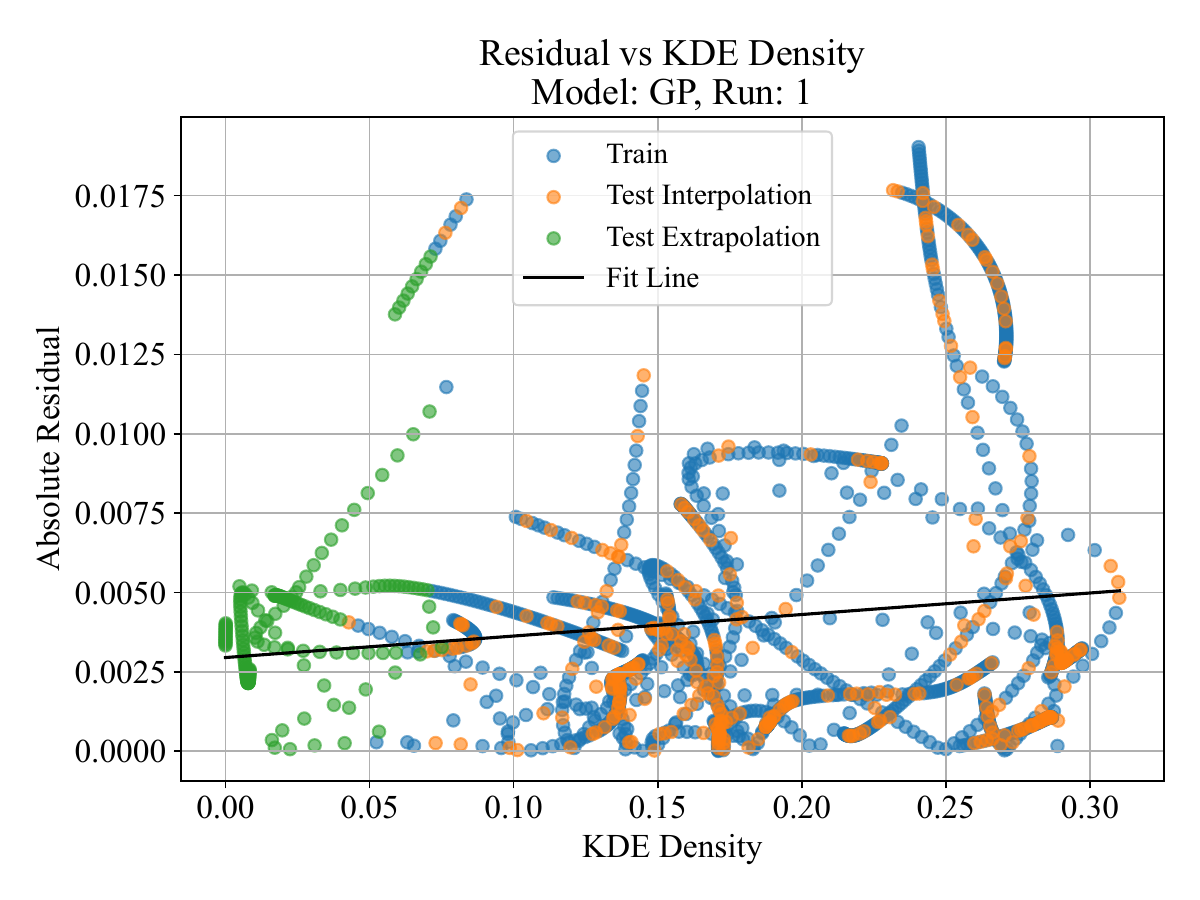}
    \caption{Absolute residuals of a GP model (Run 1) on the Gurson dataset, plotted against KDE density values. This plot shows how prediction error changes depending on how dense the data is. A slight upward trend in test extrapolation points (green) shows that error increases as data becomes more sparse.}
    \label{fig:residuals_KDE_GP}
\end{figure}

\begin{figure}[]
    \centering
    \includegraphics[width=0.9\linewidth]{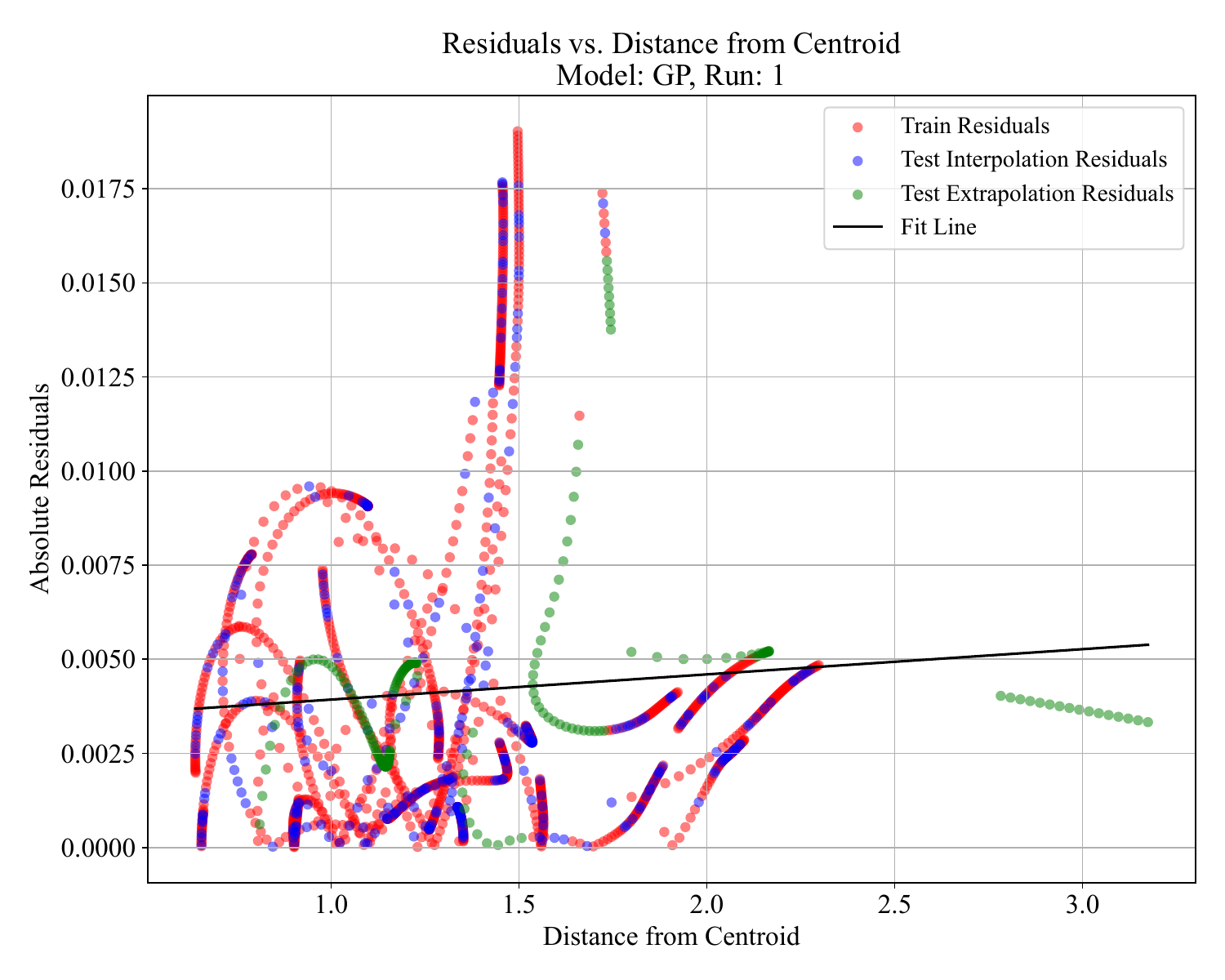}
    \caption{Absolute residuals of a GP model (Run 1) on the Gurson dataset, plotted against distance from centroid of the training data. This shows that prediction errors tend to grow the further the input is from the training data, especially in extrapolation areas.}
    \label{fig:residuals_distance_GP}
\end{figure}

\subsection{Statistical Comparison of Model Performance}\label{t-test}
To see whether the performance improvements observed from synthetic data are statistically meaningful, we performed one-sided t-tests on all datasets. 
The null hypothesis is that the synthetic data trained model does not perform better. A low p-value (here we use $p < 0.05$) shows that the improvement is statistically significant and unlikely due to chance. 

Overall, across the datasets, synthetic data led to performance improvements in many pairings, but not all of these were statistically significant. For example, in the \textbf{Gurson} dataset, out of 16 possible teacher-student combinations ($4 \times 4$), only 6 showed statistically significant improvements in extrapolation. The colour matrix shows many positive extrapolation results when synthetic data is used, especially when GP is the student model. There were 48/96 interpolation cases where the performance difference was positive, and 46 of these were statistically significant. For extrapolation, 52/96 cases had positive differences, with 43 being statistically significant. 

\subsection{Data Distribution Analysis}
Figure~\ref{fig:gurson_distribution} shows how KDE separates data into interpolation and extrapolation regions, using a sample from the Gurson dataset (Run 1). The training data is clustered in the centre area of the plot. The test interpolation data stays mostly within the same area as the training data, meaning it is similar and easier for the model to predict. However, the test extrapolation data lies farther away, in places that the model did not see during training, which makes prediction more difficult.
The extrapolation points are found in sparse parts of the feature space. This supports the idea that the KDE-based thresholding works well to separate extrapolation points from interpolation. The synthetic data, which was created by a teacher model, is nicely spread across interpolation and extrapolation areas. Some of it reaches into areas where the test extrapolation data exists, which is important. This shows that the added Gaussian noise ($\varepsilon = 0.3$) created new data points that are varied, but realistic, not too far from the original training data. This coverage helps to explain why the model performed better on extrapolation in some cases, especially for GP as the student model.

\subsection{Extrapolation Behaviour and Heterogeneous Errors}
To better understand why extrapolation is more difficult, we look at how prediction errors change depending on the location of the test data using a sample from the Gurson dataset. Figure~\ref{fig:residuals_KDE_GP} shows the absolute residuals of the GP model plotted against KDE density. A residual is the difference between the true value and the model's prediction. Lower KDE density means the data point is in a sparse area. We can see that as the KDE density decreases, the residuals (errors) increase, especially for the extrapolation data. This shows that the model struggles more when predicting in areas with little training data.

Figure~\ref{fig:residuals_distance_GP} shows the same residuals plotted against the distance from the centroid of the training data. Again, the error increases as the data points gets farther away from the training area. This pattern also affects extrapolation points the most. These two plots confirm that extrapolation is harder because of data sparsity and distance from known regions. The model's error is not the same everywhere, it changes depending on how ``familiar'' the input is.

\section{Conclusion}
This study explored how synthetic data generated by teacher models (NN, RF, GP) can be used to improve the extrapolation performance of SR models, with a particular focus on GP. We used knowledge distillation (teacher-student) framework, where the teacher model generates synthetic data in sparse (extrapolation) areas identified by kernel density estimation (KDE), then we train the same models on augmented data. KDE helps us detect under-represented or very sparse input space, allowing us to focus the synthetic data where it is most needed. Our method specifically uses synthetic data in these sparse areas, generated by teacher models (M1), to guide the training of student models. The goal is to improve the generalisation ability of the student, especially GP, in extrapolation areas where it usually struggles. The results show that this strategy often improves the extrapolation ability of GP models, especially when GP is used as the student model (M2). However, it depends on the dataset and the combination of teacher and student models.
We also found that using GP as the teacher (M1) often causes disimprovements performance in extrapolation. However, one consistent exception is when GPe is used to train GPp, this pairing always improved performance across all datasets, possibly because GPe and GPp learn in different ways that work well together.

In real world application, our method can be applied by generating synthetic data using a teacher model and training the student model on it. We can then run the trained model on a validation set to see whether it improves or worsens performance. If we observe improvement, we can expect that improvement to carry over to both the test interpolation and test extrapolation results. This makes the method more practical, flexible, and adaptable to different datasets.

\section{Acknowledgement}
This publication has emanated from research conducted with the financial support of Taighde Éireann – Research Ireland under Grant No. 18/CRT/6223.

\bibliographystyle{ACM-Reference-Format}
\bibliography{sample-base}

\appendix

\end{document}